\documentclass[sigconf]{acmart} % HRI wants us to use this for the initial submission as well
\AtBeginDocument{%
  \providecommand\BibTeX{{%
    \normalfont B\kern-0.5em{\scshape i\kern-0.25em b}\kern-0.8em\TeX}}}

\setcopyright{acmcopyright}
\copyrightyear{2024}
\acmYear{2024}
\acmDOI{XXXXXXX.XXXXXXX}

\acmConference[HRI '24]{ACM/IEEE International
Conference on Human Robot Interaction}{March 11--14,
  2024}{Boulder, CO}
\usepackage{graphicx}
\graphicspath{ {./images/} }
\usepackage{subcaption}
\usepackage{tablefootnote}

\begin{document}

% \title{When is personalized adaptation to human preferences worth it? Effects of adaptation on trust and human-robot team performance}\\
% \title{Evaluating the Impact of Personalized Adaptation in Human-Robot Interaction: Insights into Trust and Team Performance Outcomes}

\title[Impact of Personalized Value Alignment in HRI]{Evaluating the Impact of Personalized Value Alignment in Human-Robot Interaction: Insights into Trust and Team Performance Outcomes}

% improve trust on the robot and team performance?}
% worth it? \textcolor{blue}{somehow we need to point out that the DV of interest is trust and performance, otherwise your introduction paragraph 1 becomes irrelevant.}}

\author{Shreyas Bhat}
\email{shreyasb@umich.edu}
\orcid{0000-0001-6052-3178}
\affiliation{%
    \institution{University of Michigan}
    \streetaddress{1205 Beal Ave.}
    \city{Ann Arbor}
    \state{Michigan}
    \country{USA}
    \postcode{48109-2117}
}

\author{Joseph B. Lyons}
\email{joseph.lyons.6@us.af.mil}
\affiliation{%
  \institution{Air Force Research Laboratory}
  % \streetaddress{P.O. Box 1212}
  \city{Dayton}
  \state{Ohio}
  \country{USA}
  \postcode{45433}
}

\author{Cong Shi}
\email{congshi@bus.miami.edu}
\affiliation{%
  \institution{Miami Herbert Business School}
  \streetaddress{Lorem Ipsum}
  \city{Miami}
  \state{Florida}
  \country{USA}
}

\author{X. Jessie Yang}
\email{xijyang@umich.edu}
\affiliation{%
  \institution{University of Michigan}
  \streetaddress{1205 Beal Ave.}
  \city{Ann Arbor}
  \state{Michigan}
  \country{USA}  
}

\renewcommand{\shortauthors}{Bhat, et al.}

\begin{abstract}
This paper examines the effect of real-time, personalized alignment of a robot's reward function to the human's values on trust and team performance. We present and compare three distinct robot interaction strategies: a non-learner strategy where the robot presumes the human’s reward function mirrors its own; a non-adaptive-learner strategy in which the robot learns the human's reward function for trust estimation and human behavior modeling, but still optimizes its own reward function; and an adaptive-learner strategy in which the robot learns the human's reward function and adopts it as its own. 
% In all interaction strategies, trust is explicitly used in the planning subsystem of the robot. 
Two human-subject experiments with a total number of $N=54$ participants were conducted. In both experiments, the human-robot team searches for potential threats in a town. The team sequentially goes through search sites to look for threats. We model the interaction between the human and the robot as a trust-aware Markov Decision Process (trust-aware MDP) and use Bayesian Inverse Reinforcement Learning (IRL) to estimate the reward weights of the human as they interact with the robot. 
In Experiment 1, we start our learning algorithm with an informed prior of the human's values/goals. In Experiment 2, we start the learning algorithm with an uninformed prior. 
% The same informed prior is used to set the reward functions for the non-adaptive interaction strategies. 
% on the human's preferences and this prior is used to set the reward weights of the non-adaptive interaction strategies. 
Results indicate that when starting with a good informed prior, personalized value alignment does not seem to benefit trust or team performance. On the other hand, when an informed prior is unavailable, alignment to the human's values leads to high trust and higher perceived performance while maintaining the same objective team performance.
\end{abstract}

\begin{CCSXML}
<ccs2012>
   <concept>
       <concept_id>10003120.10003121.10011748</concept_id>
       <concept_desc>Human-centered computing~Empirical studies in HCI</concept_desc>
       <concept_significance>500</concept_significance>
       </concept>
   <concept>
       <concept_id>10010520.10010553.10010554.10010557</concept_id>
       <concept_desc>Computer systems organization~Robotic autonomy</concept_desc>
       <concept_significance>500</concept_significance>
       </concept>
 </ccs2012>
\end{CCSXML}

\ccsdesc[500]{Human-centered computing~Empirical studies in HCI}
\ccsdesc[500]{Computer systems organization~Robotic autonomy}
\keywords{Human-robot teaming, trust-aware decision-making, value-alignment}


% \received{29 SEP 2023}
% \received[revised]{10 NOV 2023}
% \received[accepted]{22 NOV 2023}

\maketitle

\section{Introduction}
\label{sec:intro}

Robots are increasingly becoming an integral part of our daily lives, marking their presence across varied domains, including healthcare, manufacturing, education, and home assistance, to name a few. As this integration deepens, robots are no longer perceived as tools performing isolated tasks; they are evolving as collaborative partners working with humans. In this human-robot partnership, research into trust between humans and robots becomes increasingly important \cite{sheridan_humanrobot_2016, Billings2012, Chiou2021, Yang:2017:EEU:2909824.3020230}. Without proper trust, the potential of human-robot teams remains unrealized. 

A considerable amount of research has been devoted to developing robots exhibiting trustworthy behaviors, as well as investigating methods for predicting and managing the human's trust in the robot. For instance, one specific area of research focuses on providing explanations of the robots' behaviors \cite{Wang2016, Wang:0tu, Lyons2023, lyons_explanations_2023, DU2019428}
% [citation], 
typically leading to higher perceived trustworthiness and, subsequently, trust in the robot. Other research directions include the developing real-time trust prediction algorithms \cite{Xu2015, Guo2021, guo_enabling_2023, Soh2020, yang_toward_2023}, modeling trust dynamics \cite{Cohen_dynamics_2021, yang_toward_2023, Yang2021_chapter}, developing trust repair strategies \cite{karli_what_2023, esterwood_three_2023}, and developing trust-aware planning \cite{Bhat2022, Chen2020, Kumar2019b, Chen2018, zahedi2023, Pippin2014}.
% [cite]. 

More recently, the idea of value/goal alignment -- aligning the values/goals of robots with those of humans, has garnered significant attention, with the assumption that such alignment would benefit human-robot interaction \cite{sanneman_validating_2023, Yuan2022}. Recent literature in value/goal alignment is primarily focused on enabling the autonomous or robotic agent to learn the human's values/goals through preferences \cite{Hadfield-Menell2016, biyik2018batch, christiano2023deep} or demonstrations \cite{Fisac2020, Hadfield-Menell2016, arora2021}.
% \textcolor{orange}{\cite{christiano2023deep} seems to be published in 2017 instead of 2023} \textcolor{blue}{Shreyas: @Jessie, arxiv seems to say that they last revised the paper in 2023, that is why it was citing the 2023 version. The conference (NIPS) proceedings version is from 2017. Which one should we cite? I think its better to cite the 2023 version since it is the most up-to-date.} 
% However, most existing work in value/preference alignment deals with a human-supervisor robot-worker scenario and does not explicitly incorporate the human's trust in the robot into the robot's planning and decision-making framework [e.g.,  \cite{Yuan2022}]. 
However, there is a lack of research empirically examining and quantifying the effects of alignment on human-robot interaction processes and outcomes. Yet, there are at least three reasons to suggest that such alignment could be beneficial. First, prior research has illustrated that agent adaptation to humans can enhance performance  \cite{Azevedo-Sa_2020, luo_workload_2021}.  Second, agent adaptation could be viewed as the agent being responsive to the human and may, in turn, increase human trust in the agent and enhance team performance \cite{Li2021}. Third, value alignment not only could facilitate trust establishment and enhance team performance, but it is also important for ensuring that machine partners are morally acceptable \cite{Laakasuo2023}.

This study investigates the effect of real-time, personalized alignment of a robot's reward function to the human's values on trust and human-robot team performance through two human-subject studies.
% ($N=54$) 
% where a human-robot team searches for potential threats in a town. 
We model the interaction between the human and the robot as a trust-aware Markov Decision Process (trust-aware MDP) and use Bayesian Inverse Reinforcement Learning to estimate the reward weights of the human as they interact with the robot. We compare three types of robot interaction strategies: (1) the non-learner strategy, where the robot presumes the human’s reward function mirrors its own; (2) a non-adaptive-learner strategy, in which the robot learns the human’s reward function for trust estimation and human behavior modeling, but still optimizes its own reward function; and (3) an adaptive-learner strategy where the robot learns the human’s reward function and aligns to it. In addition, we employ different initial conditions for the IRL learning algorithm in the two experiments, one with an informed prior and the other with an uninformed prior.

Results indicate that when starting with an informed prior, personalized alignment to values does not seem to benefit trust or team performance. On the other hand, when an informed prior is unavailable, aligning to the human's values leads to higher trust, agreement, and reliance intentions while maintaining the same objective team performance. To the best of our knowledge, this is one of the few studies, if not the only, that provides empirical evidence for the benefits of value alignment. 

% Results indicate that aligning the robot's values with the human's does not benefit trust or team performance when a good prior of the human's values is available. However, without a good prior, aligning with personalized values/goals leads to higher trust in the robot, higher agreement, and higher reliance intentions while maintaining team performance. 
% Further, in this uninformed prior case, although the objective performance between the three strategies is the same, subjective performance measured through the NASA TLX survey indicates that the participants feel that they have performed significantly better with the strategy that aligns values of the robot with that of the human.
% Finally, we also observe that participants tend to act somewhat riskier when interacting with the adaptive strategy compared to our non-adaptive strategies.}

The rest of the paper is organized as follows: Section \ref{sec:related-work} gives an overview of related work that our study builds upon. Section \ref{sec:formulation} details the human-robot team task and formulates our problem as a trust-aware Markov Decision Process (trust-aware MDP). Section \ref{sec:experiment} details the human-subjects experiment. Section \ref{sec:results} discusses major results and their implications. Finally, section \ref{sec:conclusion} concludes our study and discusses limitations and future work.

\section{Related Work}
\label{sec:related-work}
Our study is motivated by two bodies of research.
The first is using Inverse Reinforcement Learning (IRL) \cite{Ng2000} to learn from human demonstrations and/or preferences to guide the robot's behavior.
% and checking if the robot's reward function aligns with the values of the human. 
The second deals with trust-aware planning and quantitative modeling of trust, the goal of which is to estimate the human's trust level during interaction and to use the estimated value of trust to plan behaviors for the robot. 
\subsection{Value Alignment}
% Inverse Reinforcement Learning (IRL) \cite{Ng2000} is a field of research that deals with learning reward functions through human demonstrations, preferences, behavior, etc. Arora and Doshi \cite{arora2021} provide a comprehensive review of challenges, methods, and progress in the field of IRL.

% Ramachandran and Amir \cite{Ramachandran2007} proposed a Bayesian framework for IRL which uses the Bayes' rule over an assumed human behavior model and the evidence from experts to learn and update a distribution on the reward function. 
% \textcolor{blue}{Jessie: can we consider prior literature in IRL as having trust of 100\%? it seems to me they don't have the trust layer.}

% \textcolor{orange}{Jessie: @Shreyas, I like the way you explain all the previous work in IRL in the intro. You could consider moving some of them here} \textcolor{blue}{Done.}

Over the past few years, the problem of aligning the values/goals of the robot to those of its human teammate has been studied in detail in human-robot teaming literature \cite{Hadfield-Menell2016, Milli2017, Fisac2020, Yuan2022}. 
% In most of these works, techniques from IRL \cite{Ng2000} are used to estimate a reward function that aligns with human demonstrations, preferences, or actions. 
% \textcolor{blue}{Jessie: @Shreyas, you only describe the Yuan et al, and the Brown's study. How about the other cited work? You should at least describe them.}

A bidirectional value alignment problem is studied by Yuan et al. \cite{Yuan2022}. In their study, the human knows the true reward function and behaves accordingly while interacting as a supervisor to a group of worker robots. The robots try to learn this true reward function through correctional inputs to their behavior from the human. The human, on the other hand, tries to update her belief on the robot's belief of the true reward function and inputs corrections to their behavior accordingly. They compare the degree of alignment of human estimates of the robot's value function and the degree of alignment of the robot's value function to the true value function.
Their results reveal evidence of a bidirectional value learning behavior from the human and the robot.
% , with one's behavior being driven by the difference between one's own values and one's estimate of the others' values.

% In our case, there is no \emph{true} reward function: the human and the robot have their own reward functions, and we want to see the effect of aligning/not aligning the robot's reward function with that of the human.

% A ``driver's test" to verify value alignment between the human and the robot is provided in Brown et al. \citet{Brown2020}. 
% \citet{Brown2020} provide a ``driver's test'' to verify value alignment between the human and the robot. 
% This is especially relevant when the human and the robot are performing separate tasks in collaboration since, in this case, it is not enough to match the reward functions of the human and the robot. 
% Using their framework, a robot can check whether the learned reward function aligns with the human's values using a minimal number of queries.
% \textcolor{blue}{RESULTS of the study???}
% In our case, since the action sets for the robot and the human are the same, it is enough to match their reward functions to guarantee value alignment. 

\citet{Hadfield-Menell2016} formally define the value alignment problem as cooperative inverse reinforcement learning (CIRL). 
They show that the more traditional framework of apprenticeship learning can be formulated as a CIRL game.
% They show that finding optimal solutions to CIRL can be reduced to solving a POMDP. 
Their results indicate that the human acting optimally in isolation may not be an effective way to teach the robot. They show that under the CIRL formulation behaviors such as active teaching, active learning, and communicative actions become optimal.

\citet{Fisac2020} discusses a solution to the CIRL game based on established models of cognition and theory of mind. Under this solution, the human thinks pedagogically, choosing actions that give the most information to the robot about the underlying reward function. The robot, in turn, expects this behavior and acts pragmatically on the human's behavior. This enables the robot to learn the reward function quickly and efficiently.

\citet{christiano2023deep} proposes an algorithm to learn from human preferences and shows that it can be used to ``solve'' reinforcement learning tasks in which the robot's goal is to minimize cost to reach a goal state. They show that this can be done for complex tasks within an hour of the human's time. Additionally, they show that by incorporating human preferences, the robot can learn more efficiently than using traditional deep reinforcement learning methods.

\citet{Milli2017} compare a robot that completely abides by the human's literal order with a robot that instead behaves according to its estimate of the human's underlying preferences. 
% They use the bounded rationality model to model a noisily rational human. 
% According to there results, there should be a middle ground between complete obedience and complete reliance on the estimate of human's underlying preferences that collaborative robots should thrive to find.
They use simulations to compare how much more reward the human would get if the robot directly followed the human's orders vs if the robot used an estimate of the human's preferences. Their results indicate that 1) when a human is not rational, a robot should not directly obey their commands, 2) The optimal robot obeys only optimal commands from the human, and uses the estimate of the posterior mean on the reward features to drive its behavior otherwise.

% \textcolor{red}{Shreyas: @Jessie, I removed the Brown et al. reference and have expanded on the other references. Also removed the last difference, as I think it might not be the best thing to say here.}

There are two main
% several 
differences between our work and prior literature in value alignment. Firstly, most prior work in value alignment deals with a human-supervisor robot-worker scenario (the robot performs some task and the human is free to interrupt the task if they see any unexpected behavior from the robot) or scenarios where the human demonstrates ideal behaviors to the robot. In our case, however, we want to predict and use the probability of the human accepting or rejecting recommendations from the robot (i.e., robot-recommender human-follower scenario). This calls for the embedded trust dynamics and human behavior model, which is absent in most previous works in this area.
% Firstly, since most prior work in value alignment deals with a human-supervisor robot-worker scenario, trust was not explicitly modeled as a factor in the planning of the robot. The robot plans its behavior without the human in the loop. In our case, the human is explicitly in the loop since we are dealing with a robot-recommender human-follower scenario, and it is essential for this robot to be able to predict the accepting/rejecting behavior of the human.
Secondly, in our case, there is no \emph{true} reward function: the human and the robot have their own reward functions, and we want to see the effect of aligning/not aligning the robot's reward function with that of the human on trust and team performance. Such studies have not been done previously, according to the best of our knowledge.
% Lastly, since in our scenario, the action sets for the human and the robot are the same, we can directly adopt the human's reward function as the robot's and affirm value alignment. There is no need to ``verify'' whether their values are aligned \cite{Brown2020}.

% We use a Bayesian framework for IRL \cite{Ramachandran2007}, which learns human preferences by maintaining and updating a distribution over the possible preferences of the human. The update happens in a Bayesian way after observing the human's selected action. 

\subsection{Trust-Aware Decision Making}
In recent years, there has been substantial research in developing trust-aware decision-making algorithms for robots that work with humans in collaborative tasks. These works model the human's trust in the robot explicitly in the decision-making framework and leverage the use of quantitative trust and trust-behavior models.

Guo et al. \cite{Guo2021b} and Bhat et al. \cite{Bhat2022} present the use of the Beta distribution trust model \cite{Guo2021} in a sequential decision-making scenario for a human-robot team. They use this model to define a trust-aware Markov Decision Process, which gives optimal actions for the robot considering the human's trust and subsequent behavior. 

% \textcolor{blue}{Jessie: @Shreyas, add Yaohui's latest work about TIP. And also Harold Soh's work of using Gaussian Process to estimate trust.}

Akash et al. \cite{Kumar2019a} model the human-robot interaction as a Partially Observable Markov Decision Process (POMDP) with human's trust and workload as the states. A solution to their formulation is presented in \cite{Kumar2019b} which gives optimal level of transparency for the robot's interface depending on the human's level of trust. 

Chen et al. \cite{Chen2020} propose a trust-POMDP that is solved to generate optimal trust-based policies for the robot. They demonstrate it in a human-subject experiment involving pick-and-place tasks for a human-robotic arm team. The robot chooses an ``easy" object to pick and place to build trust and moves to harder objects when trust is high to minimize interruptions from the human. 

Zahedi et al. \cite{zahedi2023} present a trust dynamics model and a meta-MDP framework that chooses a robot's behavior depending on the level of trust of the human. They analyze the case where the human's model of the environment may be false leading to sub-optimal trustworthy policies and untrustworhy optimal policies. Their framework generates an optimal policy for the robot that chooses the trustworthy sub-optimal policies when trust is low in order to increase it and chooses the untrustworthy optimal policy when trust is high enough to improve the team's performance.

A majority of these works model the human-robot interaction as a reward-maximization problem with a reward function that is known to the team. Our work differs in this respect; we offer participants the autonomy to formulate their own reward functions, providing them only with a broad understanding of the team's objectives. Subsequently, we explore how discrepancies in reward functions between humans and robots influence trust and overall team performance.

% \subsection{Value Alignment}

\section{Problem Formulation}
\label{sec:formulation}
This section describes the task for the human-robot team and the mathematical formulation of the interaction. 
\subsection{Human-Robot Teaming Task}
We designed a scenario in which the human-robot team performs a search for potential threats in a town. The team sequentially goes through search sites to look for threats. At each site, the team is given a probability of threat presence inside the site via a scan of the site by a drone. The robot additionally, has some prior information about the probability of threat presence at all of the search sites. This prior information is unknown to the human, thus creating interdependence between the human and drone. After getting the updated probability of threat presence, the robot solves the trust-aware MDP to generate a recommendation for the human. It can either recommend the human to use or not use an armored robot for protection from threats. Encountering a threat without protection from the armored robot will result in injury to the human. On the other hand, using the armored robot takes extra time since it takes some time to deploy and move the armored robot to the search site. The goal of the team is to finish the search mission as quickly as possible while also maintaining the soldier's health level. Thus, a two-fold objective arises with conflicting sub-goals: To save time you must take risks, and if you want to avoid risks, you must sacrifice precious mission time. 

\subsection{Markov Decision Process}
\label{sec:MDP}
We model the interaction between the human and the robot as a trust-aware Markov Decision Process (trust-aware MDP), which is 
% . A trust-aware MDP is 
a tuple of the form $(S, A, T, R, H)$, where $S$ is a set of states one of which is the trust of the human on the robot, $A$ is a finite set of actions, $T$ is the transition function, 
% giving the transition probabilities from one state to another given an action, 
$R$ is a reward function and $H$ is an embedded human trust-behavior model, which gives the probabilities of the human choosing a certain action given the action chosen by the robot, their level of trust, etc. Below we provide details of our MDP formulation. 
\subsubsection{States}
We use the level of trust $t \in [0, 1]$ as the state in our trust-aware MDP formulation. The dynamics of trust are thus described by our transition function.
\subsubsection{Actions}
At any site, the recommender robot has two choices of action. It can either recommend to use or not use the armored robot. These are represented by the binary actions $a^r=1$ and $a^r=0$, respectively. Thus, our action set is $A=\{0,1\}$. After receiving a recommendation, the human chooses action $a^h$ from the same action set. 
\subsubsection{Reward Function}
The rewards for both agents (the human and the robot) are a weighted sum of the negative cost of losing health and losing time. The weights for these costs can be different for the robot and the human. In general, for agent $o\in\{h,r\}$, the reward function can be written as,
\begin{equation}
    \label{eq:reward-function}
    R^o(D, a) = -w^o_h h(D, a) - w^o_c c(a).
\end{equation}
Here, $D$ is a random variable representing the presence of threat inside a search site, $a$ is the action chosen by the human to implement, $o\in\{h,r\}$ represents the agent, either the human $h$ or the robot $r$. Note that $h(D, a)$ is a function giving the health loss cost and $c(a)$ is a function giving the time loss cost. 

\subsubsection{Transition Function}
The transition function gives the dynamics of trust as the human interacts with the robot. We use the model from \citet{Bhat2022} which models trust as a random variable following the Beta distribution based on personalized parameters $(\alpha_0, \beta_0, v^s, v^f)$.  Here, $\alpha_0, \beta_0$ are a measure of the initial trust of the human, while $v^s$ and $v^f$ control the effect of successes and failures on trust respectively. More specifically, the trust level after $i$ interactions is given by
\begin{equation}
    \label{eq:trust-update}
    t_i \sim Beta(\alpha_i, \beta_i),
\end{equation} 
where the parameters are updated through
\begin{align}
    \alpha_i &= \alpha_0 + \sum_{j=1}^ip_jv^s,\\
    \beta_i &= \beta_0 + \sum_{j=1}^i(1-p_j)v^f.
\end{align}
Here, $i$ is the number of interactions completed between the human and the robot, $t_i$ is the current level of trust, and $p_j$ is the realization of the random variable performance $(P_j)$ of the recommender robot at the $j^{th}$ interaction. It is defined below.
\begin{equation}
\label{eq:performance}
    P_j =
    \begin{cases}
        1, ~\text{ if } R^h_j(a^r_j) \geq R^h_j(1-a^r_j),\\
        0, ~\text{ otherwise.}
    \end{cases}
\end{equation}
Here, $R^h_j(a^r_j)$ is the reward for the human for choosing the recommended action $(a^r_j)$ at the $j^{th}$ interaction and $R^h(1-a^r_j)$ is the reward for choosing the other action $(1-a^r_j)$. Thus, essentially, the transition function shifts the distribution to the right if the recommend action led to a better immediate reward to the human. Otherwise, it shifts the distribution to the left. 

\subsubsection{Human Trust-Behavior Model}
\label{sec:behavior-model}
A human trust-behavior model gives the probabilities of a human choosing an action, given the robot's action, their trust level, and other factors such as the human's goals/values. In our study, we make use of the Bounded Rationality Disuse Model of human trust-behavior. This model states that the human chooses the recommended action with a probability equal to the human's current level of trust. If the human chooses to ignore the recommendation, s/he will choose an action according to the bounded rationality model of human behavior. That is, the human will choose an action with a probability that is proportional to the exponential of the expected reward the human receives with that action. Mathematically,
\begin{align}
\label{eq:behavior-model}
    P(a^h_i=a|a^r_i=a) &= t_i + (1-t_i)q_a,\\
    P(a^h_i=1-a|a^r_i=a) &= (1-t_i)(1-q_a).
\end{align}
where $t_i$ is the human's level of trust at the $i^{th}$ interaction and $q_a$ is the probability of choosing action $a\in\{0,1\}$ under the bounded rationality model \cite{Milli2017, Ziebart2010, Fisac2020}. It is given by,
\begin{equation}
    q_a = \frac{\exp(\kappa E[R^h_i(a)])}{\sum_{a'\in\{0,1\}}\exp(\kappa E[R^h_i(a')])}.
\end{equation}
Here, $\kappa$ is called the rationality coefficient of the human, with a higher value indicating a more rational human, and a value of $0$ indicating a human that chooses an action at random.
Note that this model can be easily extended to the case where multiple actions are possible for the human-robot team. We will just need to sum over all actions in the denominator to get the probabilities.
% This is one of the advantages of using this model over the previously used ``Reverse Psychology Model \cite{Bhat2022, Guo2021b}''.

\subsection{Bayesian Inverse Reinforcement Learning}
\label{sec:Bayes-IRL}
We use Bayesian IRL to estimate the reward weights of the human as they interact with the recommender robot. This is done by maintaining a distribution on the possible reward weights and updating it using Bayes' rule after observing the human's selected action. More precisely, if $b_i(w)$ is the belief distribution on the reward weights before the $i^{th}$ interaction, the distribution after the $i^{th}$ interaction, $b_{i+1}(w)$ is given by,
\begin{equation}
\label{eq:Bayes-IRL}
    b_{i+1}(w) \propto 
    \begin{cases}
        P(a^h_i=a^r_i|a^r_i)b_i(w), \phantom{1-a}\text{ if } a^h_i = a^r_i,\\
        P(a^h_i=1-a^r_i|a^r_i)b_i(w), ~\text{ otherwise.}
    \end{cases}
\end{equation}

In our formulation, we only learn a distribution over the health reward weight of the human, $w^h_h$, and assume that the time reward weight is defined by $w_c^h := 1-w^h_h$. We use the mean of the learnt distribution as an estimate of the human's health reward weight.
% the mean of the distribution acts as a proxy for the human's health reward weight estimate. 
The Bayesian IRL algorithm requires a prior distribution $b_0(w)$ to get started. We ran the algorithm starting with a uniform prior on previously collected data \cite{Bhat2022} and generated an ``average" distribution that can represent the weights for the general population. In Experiment $1$, we use this informed prior for $b_0(w)$ and in Experiment $2$, we use the uniform distribution for $b_0(w)$ to simulate a case where we lack data to set an informed prior. 

\section{Experiment}
\label{sec:experiment}
This section provides details about the testbed and the human-subject experiments. The experiments complied with the American Psychological Association code of ethics and were approved by the Institutional Review Board at the University of Michigan.

\subsection{Testbed}
\label{sec:testbed}

% \textcolor{blue}{Jessie: @Shreyas, we should make this paper self-sufficient instead of referring the participants to our previous publication. You can directly copy-paste from the previous study and paraprhase.}
% \textcolor{red}{Done.}

\begin{figure}[h]
    \centering
    \includegraphics[width=0.75\linewidth]{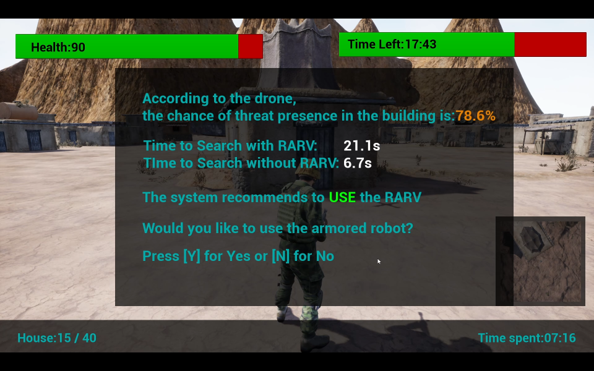}
    \caption{The recommendation interface}
    \label{fig:rec-interface}
\end{figure}

\begin{figure}[h]
    \centering
    \includegraphics[width=0.75\linewidth]{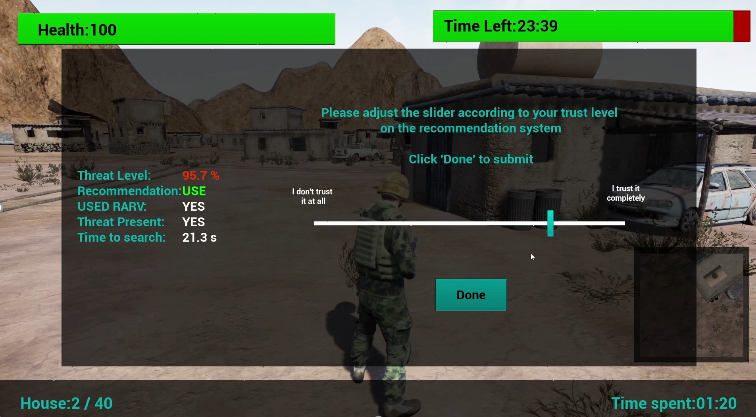}
    \caption{The trust feedback slider used to get feedback from the participants after every search site. The mission timer is paused when the slider is shown to let the participants take their time in adjusting their trust.}
    \label{fig:feedback-slider}
\end{figure}

We developed a 3D testbed using the Unreal Engine game development platform. A soldier moves with an autonomous drone in a town to search for threats (armed gunmen). Before entering a site, the drone scans the site and reports the chance of threat presence inside the site (Fig. \ref{fig:rec-interface}). Then, the participant is presented with the average time taken to search a site with and without the armored robot to aid their decision. Finally, the participants are given the recommendation generated by the intelligent agent. 
% Once the participant makes their decision, they are first shown their chosen action. 
If the participant chooses to use the armored robot, the armored robot is shown to be moving towards the site. This deployment of the armored robot takes about $15$ seconds. If the participant chooses to not use the armored robot, they enter the site directly without any time loss. In case a threat is encountered without protection from the armored robot, the participant loses $5$ points of health. The participant does not lose any health if there is no threat or if there is a threat but the participant chose to use the armored robot. 
% They are instructed to choose an action based on their interaction history and the recommendation from the intelligent agent. 
After exiting each house, the participants are asked to adjust a slider to give feedback on their level of trust on the agent's recommendations (Fig. \ref{fig:feedback-slider}). The feedback slider shows the threat level, the recommendation, the participant's choice, the presence of threat, and the time it took to search the site to help the participants in assessing their 
% level of 
trust. 

% We updated some elements of the testbed from \cite{Bhat2022} for this study, following feedback from participants in that study. The testbed was developed in the Unreal Engine game development platform. We updated the recommendation interface to be more informative and to help the participants make their own decisions if they choose to do so. In the updated interface (shown in fig. \ref{fig:rec-interface}), the participants are shown the probability of threat presence reported by the drone, the recommendation given by the intelligent agent, and a rough estimate of search time with and without the armored robot. 

% Further, we updated the trust feedback slider to provide information about the last interaction that the participant had, in order to help them make an informed decision about their trust. The updated interface can be seen in fig. \ref{fig:feedback-slider}. 

\subsection{Participants}
We collected data from a total of $54$ participants, $30$ of which participated in experiment $1$ (Age: Mean $22.6$ years, \textit{SD} $3.6$, 14 Female) and $24$ participated in experiment $2$ (Age: Mean $21.4$ years, \textit{SD} $2.3$, 12 Female). 
% Table \ref{tab:demographics} presents the demographics of the participants. 
All participants were students from the University of Michgan. 
% The study was approved by the Institutional Review Board of [Anonymity]. % Anonymization for peer review.
% All data points considered outliers $(Mean \pm 3 SD)$ were removed before performing the following data analysis.
% from the college of engineering at the University of Michigan. 

% \begin{table}[h]
%     \centering
%     \begin{tabular}{|c|c|c|c|}
%     \hline
%        Experiment & $N$ & Age (Mean $\pm$ SD) & Gender\\
%     \hline
%         1 & 30 & 22.6 $\pm$ 3.6 years & $14$ Female, $16$ Male\\
%         2 & 24 & 21.4 $\pm$ 2.3 years & $12$ Female, $12$ Male\\
%     \hline
%     \end{tabular}
%     \caption{Participant demographics}
%     \label{tab:demographics}
% \end{table}

% \subsection{Interaction Strategies and Design of Experiment}
\subsection{Experiment Design}
\label{sec:strategies}
We designed three interaction strategies for the intelligent agent:
\begin{itemize}
    \item \textbf{Non-learner:} The intelligent agent does not learn the reward weights of the human. It assumes that the human and the intelligent agent share the same reward weights and uses these for recommendation success assessment, trust updating, human behavior modeling, and MDP optimization. 
    \item \textbf{Non-adaptive learner:} The intelligent agent learns personalized reward weights for each human it interacts with. It only uses these learned weights for recommendation success assessment, trust updating, and human behavior modeling. It still optimizes the MDP with its own fixed reward weights. 
    \item \textbf{Adaptive learner:} The intelligent agent learns personalized reward weights for each human. It uses them for recommendation success assessment, trust updating, and human behavior modeling, and also optimizes the MDP based on these reward weights. 
    % In other words, it updates its own reward function to match the learned reward function. 
\end{itemize}

We employed a within-subjects design. Each participant completed three missions. In each mission, they interacted with an intelligent agent using one of the interaction strategies. 
To minimize potential order effects, a $3 \times 3$ Latin Square design was used.

% We went with a within-subjects design for the study. The reason behind this was that our goal was to compare trust between different interaction strategies. Given the high variation between trust dynamics between participants \cite{Bhat2022}, we think that it is better to compare trust within-subjects than between-subjects. The participants completed 3 missions in total. In each mission, they interacted with an intelligent agent following one of the interaction strategies (detailed in sec. \ref{sec:strategies}). In each mission, they sequentially searched through 40 search sites with the intelligent agent. 

\subsection{Measures}
% \textcolor{red}{Should we add all measures that we used here? We are not using a lot of the measures in our data analysis for this paper. For example, personality, trust propensity, demographics, rationality, type of decision-making style, etc. We are not using a majority of the pre-experiment measures.}

% \textcolor{blue}{Jessie: As we are not using them, let's simplify them}. 
\subsubsection{Pre-experiment Measures}
% We used the following measures as a part of a pre-experiment survey for the participants.
Prior to the experiment, participants filled in a demographic survey indicating their age, gender, academic department, nationality, frequency and skill of playing video games, and familiarity with AI/ML algorithms. Participants also filled in questionnaires about their personality, propensity to trust autonomy, and decision-making style. 

\subsubsection{Pre-mission Measures}Before each of the three missions, participants rated their preferences. 
% We used the following measures as a part of a pre-experiment survey for the participants.
\begin{itemize}
    \item \textbf{Task Preference:} Before the beginning of each mission, we ask the participants to rate their preference between saving health and saving time by moving a slider between these two objectives, showing their relative importance.
\end{itemize}

\subsubsection{In Experiment Measures}
After each site's search was completed (i.e., every trial), the participants were asked to report their level of trust in the intelligent agent, $t_{i}$ (see fig. \ref{fig:feedback-slider} for the exact question asked during the interaction). The slider values were between 0 and 100 with a step of 2 points. Additionally, for every trial, we measured whether the participants agreed with the intelligent agent. With the trial-based data, we measured the following:

\begin{itemize}
    \item \textbf{Average Trust:} This was calculated as the empirical mean trust $\frac{1}{M}\sum_{i=1}^M t_i$.
    \item \textbf{End-of-mission Trust:} This was the participant's self-reported trust after the last trial, $t_M$.
   \item \textbf{Number of Agreement:} This was computed as the number of times the participant chose the recommended action.
    \end{itemize}
  Note that $M=40$ is the number of sites in a mission.

\subsubsection{Post-mission Measures} After every mission, participants filled out a post-mission survey gauging the following items. 
\begin{itemize}
    \item \textbf{Post-mission trust questionnaire:} This was measured using Muir's trust questionnaire \cite{Muir1996}. It has 9 questions, each with a slider range between 0 and 100. 
    \item \textbf{Post-mission Reliance Intentions:} This was measured using the scale developed in \citet{Lyons2019}. We used 6 of the 10 items that were relevant for this task. Each item was rated on a 7-point Likert scale.
    \item \textbf{Workload:} Workload was measured using the NASA TLX scale \cite{Hart1988}. We used 5 of the 6 dimensions as our experiment involved minimal physical demand. Each item was measured using a slider ranging from very low to very high. 
    % We calculated the average workload by taking the average over the ratings on the 5 dimensions. The scale for perceived performance was reverse coded.
    \item \textbf{Performance:} We computed the team performance by a weighted sum of the percentage health remaining of the soldier and the percentage time remaining in the mission. 

   \begin{equation}
       \text{Performance} = \hat{w}^h_h\cdot (\% h_M) + \hat{w}^h_c \cdot (100 - \%c_M).
   \end{equation}
   
   % \textcolor{blue}{Jessie: @Shreyas, write down the formula here}

    where $\hat{w}^h_h$ and $\hat{w}^h_c:=1-\hat{w}^h_h$ are the reported preferences by the participant before beginning the mission, $\%h_M$ is the percent health remaining and $\%c_M$ is the percent time spent at the end of the mission. This metric allows us to convert the two conflicting objectives with different units of measurement into one unified scale. The higher the value, the better the team performance. 
    
    % Performance was measured using three metrics: the end-of-mission health level of the soldier and the total time taken to complete the mission. 
    % , and the rewards earned by the human, by the robot, and by the team as a whole.
\end{itemize}

% \subsection{Experimental Procedure}

\section{Results}
\label{sec:results}
This section summarizes our results and discusses the implications. Table \ref{tab:results} tabulates the results from the two experiments. 
Repeated measures analyses of variance (ANOVAs) were conducted to compare the three interaction strategies. Greenhouse-Geisser corrections to the degrees of freedom were made whenever a measure failed Mauchly's test of sphericity. In Experiment 1, we initiated the IRL learning algorithm with an informed prior. In Experiment 2, the learning algorithm started with an uninformed prior. 

\begin{table*}[ht]
    \centering
    \caption{Mean and standard deviation (SD) of dependent measures for the three interaction strategies in Experiments 1 and 2}
    \label{tab:results}
    \begin{tabular}{|lccc|}
\hline
\multicolumn{4}{|c|}{\textbf{Experiment $1$: with informed prior}}                                                                                                                                      \\ \hline
\multicolumn{1}{|l|}{}                                           & \multicolumn{1}{c|}{\textbf{Non-learner (Mean$\pm$SD)}}       & \multicolumn{1}{c|}{\textbf{Non-adaptive learner (Mean$\pm$SD)}} & \textbf{Adaptive-learner (Mean$\pm$SD)}  \\ \hline
\multicolumn{1}{|l|}{Average trust $\frac{1}{M}\sum_{i=1}^Mt_i$} & \multicolumn{1}{c|}{$0.73 \pm 0.16$}   & \multicolumn{1}{c|}{$0.71 \pm 0.20$}      & $0.72 \pm 0.14$   \\
\multicolumn{1}{|l|}{End-of-mission trust $t_M$}                   & \multicolumn{1}{c|}{$0.81\pm 0.18$}    & \multicolumn{1}{c|}{$0.76 \pm 0.21$}      & $0.76 \pm 0.16$   \\
\multicolumn{1}{|l|}{Muir's trust questionnaire}                 & \multicolumn{1}{c|}{$74.20\pm 14.80$}  & \multicolumn{1}{c|}{$71.09 \pm 23.17$}    & $68.28 \pm 17.4$  \\
\multicolumn{1}{|l|}{Number of agreements}                       & \multicolumn{1}{c|}{$36.03\pm 3.72$}   & \multicolumn{1}{c|}{$35.27 \pm 4.68$}     & $36.00 \pm 2.60$  \\
\multicolumn{1}{|l|}{Reliance intentions scale $^\ast$}                  & \multicolumn{1}{c|}{$4.34 \pm 1.01$}   & \multicolumn{1}{c|}{$4.10 \pm 1.35$}      & $3.70 \pm 1.27$   \\
\multicolumn{1}{|l|}{Workload}                                   & \multicolumn{1}{c|}{$36.87\pm 18.04$}  & \multicolumn{1}{c|}{$34.45\pm 17.02$}     & $39.03\pm 16.85$  \\
\multicolumn{1}{|l|}{Performance}                                & \multicolumn{1}{c|}{$61.47 \pm 18.12$} & \multicolumn{1}{c|}{$60.25\pm 17.03$}     & $55.83\pm 20.39$  \\ \hline
\multicolumn{4}{|c|}{\textbf{Experiment $2$: with uninformed prior}}                                                                                                                                      \\ \hline
\multicolumn{1}{|l|}{}                                           & \multicolumn{1}{c|}{\textbf{Non-learner (Mean$\pm$SD)}}       & \multicolumn{1}{c|}{\textbf{Non-adaptive learner (Mean$\pm$SD)}} & \textbf{Adaptive-learner (Mean$\pm$SD)}  \\ \hline
\multicolumn{1}{|l|}{Average trust $\frac{1}{M}\sum_{i=1}^Mt_i$ $^\ast$} & \multicolumn{1}{c|}{$0.42 \pm 0.20$}   & \multicolumn{1}{c|}{$0.45 \pm 0.22$}      & $0.65 \pm 0.20$   \\
\multicolumn{1}{|l|}{End-of-mission trust $t_M$ $^\ast$}                   & \multicolumn{1}{c|}{$0.33 \pm 0.30$}   & \multicolumn{1}{c|}{$0.35 \pm 0.29$}      & $0.64 \pm 0.30$   \\
\multicolumn{1}{|l|}{Muir's trust questionnaire $^\ast$}                 & \multicolumn{1}{c|}{$34.51\pm 19.79$}  & \multicolumn{1}{c|}{$34.97 \pm 17.41$}    & $60.57\pm 23.71$  \\
\multicolumn{1}{|l|}{Number of agreements $^\ast$}                       & \multicolumn{1}{c|}{$28.17 \pm 5.49$}  & \multicolumn{1}{c|}{$28.67 \pm 4.65$}     & $35.62 \pm 2.93$  \\

\multicolumn{1}{|l|}{Reliance intentions scale $^\ast$}                  & \multicolumn{1}{c|}{$2.16 \pm 1.22$}   & \multicolumn{1}{c|}{$2.38 \pm 1.16$}      & $3.58 \pm 1.32$   \\
\multicolumn{1}{|l|}{Workload $^\ast$}                                   & \multicolumn{1}{c|}{$38.18 \pm 13.79$} & \multicolumn{1}{c|}{$39.82\pm 14.94$}     & $31.13 \pm 10.64$ \\
\multicolumn{1}{|l|}{Performance}                                & \multicolumn{1}{c|}{$45.86 \pm 16.20$} & \multicolumn{1}{c|}{$49.11 \pm 14.25$}    & $51.60 \pm 18.68$ \\ \hline
\multicolumn{4}{l}{\footnotesize $\ast - p<0.05$}
\end{tabular}
\end{table*}

\subsection{Trust, Agreement, and Reliance Intention }
% We expect that aligning with human preferences will result in higher levels of trust of the human on the robot. 

\subsubsection{Experiment 1: with informed prior}
% Fig. \ref{fig:trust-comparison-2a} shows the mean and standard error (SE) of (a) the average trust, (b) the end-of-mission trust, and (c) the number of times the participant's decision agreed with the recommendation across the three interaction strategies.

We observed no significant difference between the three strategies in average trust $(F(2, 58) = 0.308,~ p = 0.736)$, end-of-mission trust $(F(2, 58) = 1.192,~ p = 0.311)$, and the  Muir's trust scale ($F(2, 58) = 1.550, p = 0.221$).

% \begin{figure}[h]
%     \centering
%     \includegraphics[width=0.8\columnwidth]{images/barcharts/phase2a/trust/muir.png}
%     \caption{Post-condition subjective trust ratings}
%     \label{fig:trust-muir-2a}
% \end{figure}

% \begin{figure*}[ht]
%     \centering
%     \begin{subfigure}[t]{0.3\textwidth}
%         \centering
%         \label{fig:avg-trust-2a}
%         \includegraphics[width=\textwidth]{images/barcharts/phase2a/trust/average.png}
%         \caption{Average trust}
%     \end{subfigure}\hfil
%     \begin{subfigure}[t]{0.3\textwidth}
%         \centering
%         \label{fig:end-trust-2a}
%         \includegraphics[width=\textwidth]{images/barcharts/phase2a/trust/end-of-mission.png}
%         \caption{End-of-mission trust}
%     \end{subfigure}\hfil
%     \begin{subfigure}[t]{0.3\textwidth}
%         \centering
%         \label{fig:agreements-2a}
%         \includegraphics[width=\textwidth]{images/barcharts/phase2a/trust/number-of-agreements.png}
%         \caption{Number of agreements}
%     \end{subfigure}
%     \caption{Comparison of reported trust between the three interaction strategies. (Number of agreements is the number of times the participant selected the recommended action.)\textcolor{blue}{In the captions, please show the equation for calculating the metric, for example, End-of-mission trust $t_{40}$}}
%     \label{fig:trust-comparison-2a}
% \end{figure*}

\begin{figure}[h]
    \centering
    \includegraphics[width=0.7\columnwidth]{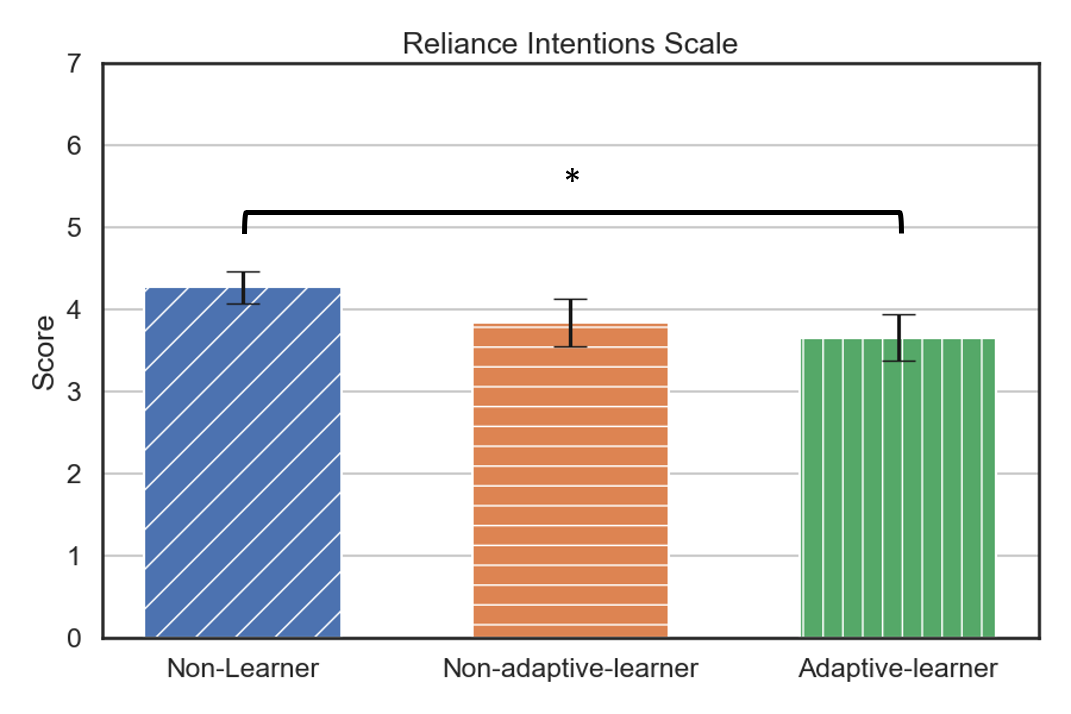}
    \caption{Exp 1 -- Post-mission reliance intentions}
    \label{fig:reliance-intentions-2a}
\end{figure}

% Fig. \ref{fig:trust-muir-2a} shows the post-mission trust rating given using Muir's trust scale. 
Additionally, there was no significant difference in the number of agreements $(F(2, 58) = 0.755,~ p = 0.475)$ across the three strategies. However, there was a significant difference in reliance intentions $(F(1.543, 44.737), ~p=0.031)$
% ($F(2, 58) = 4.171,~ p = 0.020$) 
(Fig. \ref{fig:reliance-intentions-2a}). Pairwise comparisons with Bonferroni adjustments revealed a lower intent to rely on the adaptive learner strategy than the non-learner strategy $(p=0.012)$. 

% Fig.\ref{fig:reliance-intentions-2a} shows the post mission reliance intentions rating \cite{Lyons2020} given by the participants. 

\subsubsection{Experiment 2: with uninformed prior}

% Fig. \ref{fig:trust-comparison-2b} shows the comparison of (a) average trust rating given by the participant to the recommender robot during the mission, (b) the end-of-mission trust rating given by the participant (i.e. the trust feedback given by the participant after the last site was searched), and (c) the number of times the participant's decision agreed with the recommendation.

\begin{figure}[h]
    \centering
    \includegraphics[width=0.7\columnwidth]{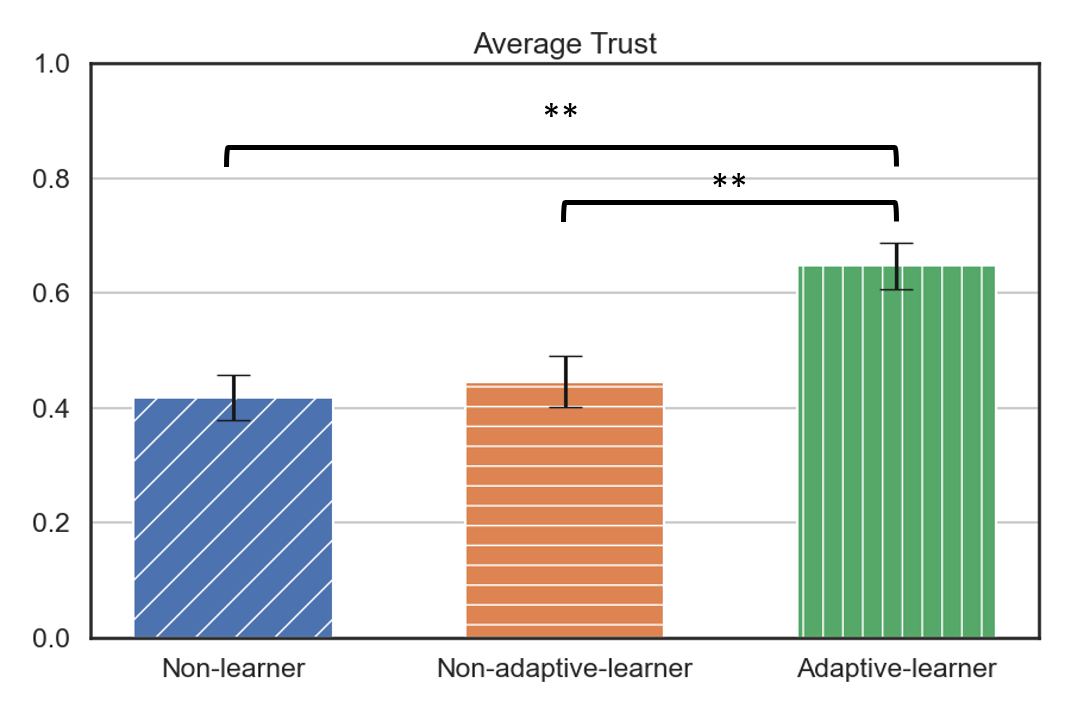}
    \caption{Exp 2-- Average trust $\frac{1}{M}\sum_{i=1}^Mt_i$}
    \label{fig:avg-trust-2b}
\end{figure}

\begin{figure}[h]
    \centering
    \includegraphics[width=0.7\columnwidth]{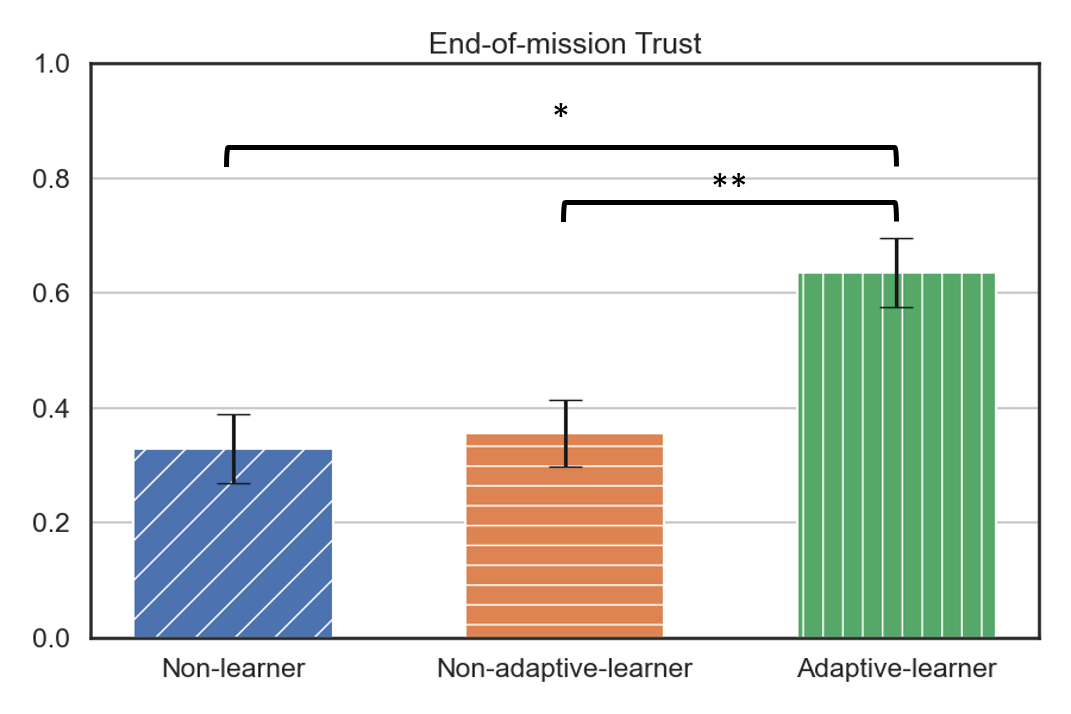}
    \caption{Exp 2-- End-of-mission trust $t_M$}
    \label{fig:end-trust-2b}
\end{figure}

\begin{figure}[h]
    \centering
    \includegraphics[width=0.7\columnwidth]{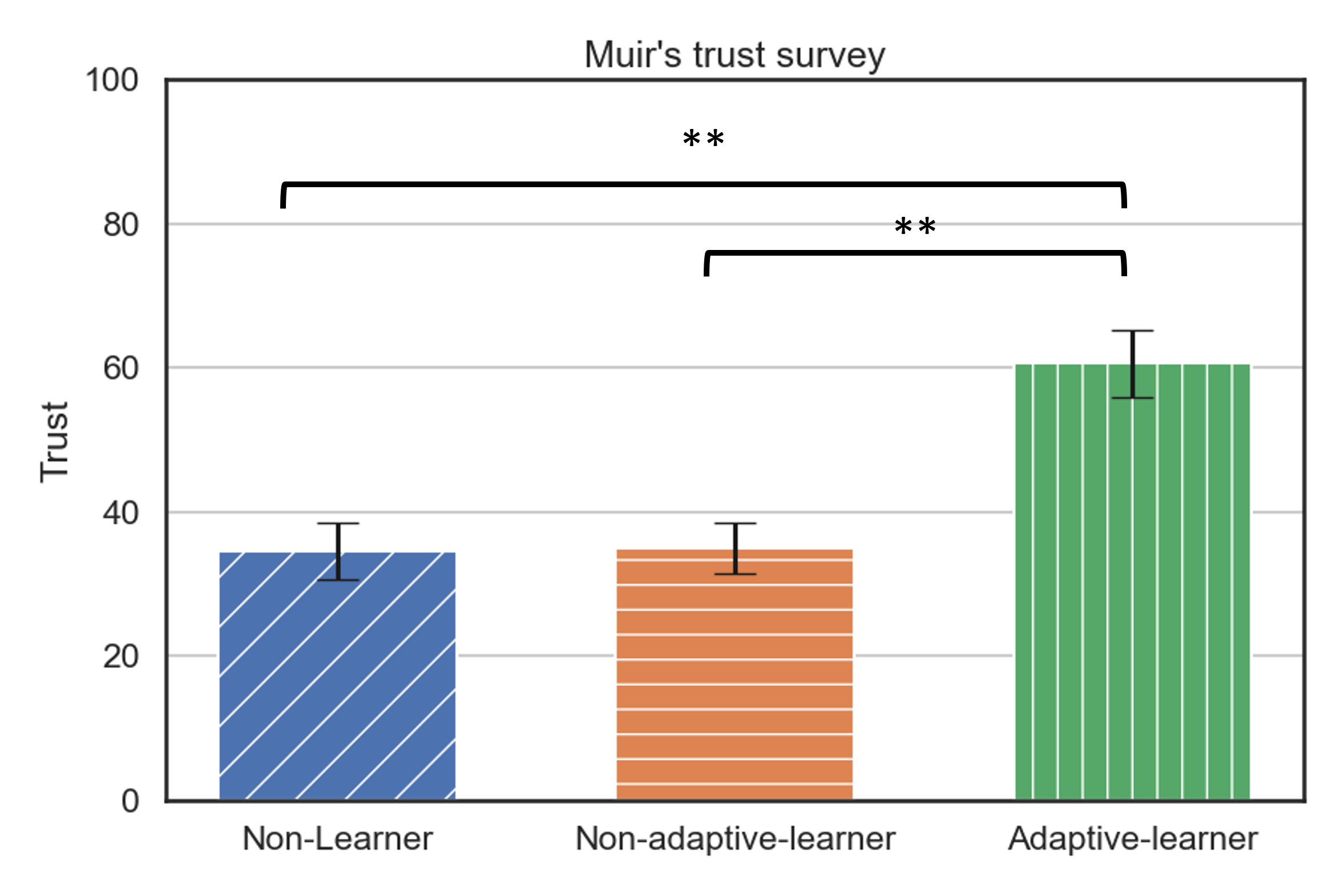}
    \caption{Exp 2-- Post-mission trust questionnaire}
    \label{fig:trust-muir-2b}
\end{figure}

Figs. \ref{fig:avg-trust-2b}, \ref{fig:end-trust-2b}, \ref{fig:trust-muir-2b} show the comparisons of the three strategies in trust. Repeated measures ANOVA revealed significant differences between the three strategies in average trust $(F(2, 46)=14.161, ~p < 0.001)$, end-of-mission trust $(F(2,46))=12.736, ~p<0.001$, and Muir's trust scale $(F(1.586,\\ 36.473) = 16.3, ~p < 0.001)$.
% ($F(2,46) = 16.3, p < 0.001$).  
Pairwise comparisons with Bonferroni adjustments revealed that the adaptive-learner strategy led to higher average trust, end-of-mission trust, and post-mission trust compared to the non-learner strategy ($p<0.001$, $p=0.001$ and $p<0.001$, respectively) and compared to the non-adaptive learner strategy ($p=0.003$, $p<0.001$, $p<0.001$, respectively). 

% significant difference between the non-learner and the adaptive-learner strategies in average trust $(p<0.001)$ and end-of-mission trust $(p=0.001)$ and between the non-adaptive-learner and the adaptive-learner strategies in average trust $(p0.003)$ and end-of-mission trust $(p<0.001)$. These differences can be seen in Fig. \ref{fig:trust-comparison-2b}. Additionally, there are significant differences between the non-learner strategy and the adaptive-learner strategy ($p<0.001$) and between the non-adaptive-learner strategy and the adaptive-learner strategy ($p<0.001$) in Muir's trust scale (Fig. \ref{fig:trust-muir-2b}). 

% \vspace{-10pt}

% \vspace{-10pt}

% \begin{figure*}[ht]
%     \centering
%     \begin{subfigure}[t]{0.3\textwidth}
%         \centering
%         \label{fig:avg-trust-2b}
%         \includegraphics[width=\textwidth]{images/barcharts/phase2b/trust/average.png}
%         \caption{Average trust}
%     \end{subfigure}\hfil
%     \begin{subfigure}[t]{0.3\textwidth}
%         \centering
%         \label{fig:end-trust-2b}
%         \includegraphics[width=\textwidth]{images/barcharts/phase2b/trust/end-of-mission.png}
%         \caption{End-of-mission trust}
%     \end{subfigure}\hfil
%     \begin{subfigure}[t]{0.3\textwidth}
%         \centering
%         \label{fig:agreements-2b}
%         \includegraphics[width=\textwidth]{images/barcharts/phase2b/trust/muir-trust.png}
%         \caption{Post-mission trust questionnaire}
%     \end{subfigure}
%     \caption{Experiment 2: Comparison of reported trust between the three interaction strategies. (Number of agreements is the number of times the participant selected the recommended action.)}
%     \label{fig:trust-comparison-2b}
% \end{figure*}

\begin{figure}[h]
    \centering
    \includegraphics[width=0.7\columnwidth]{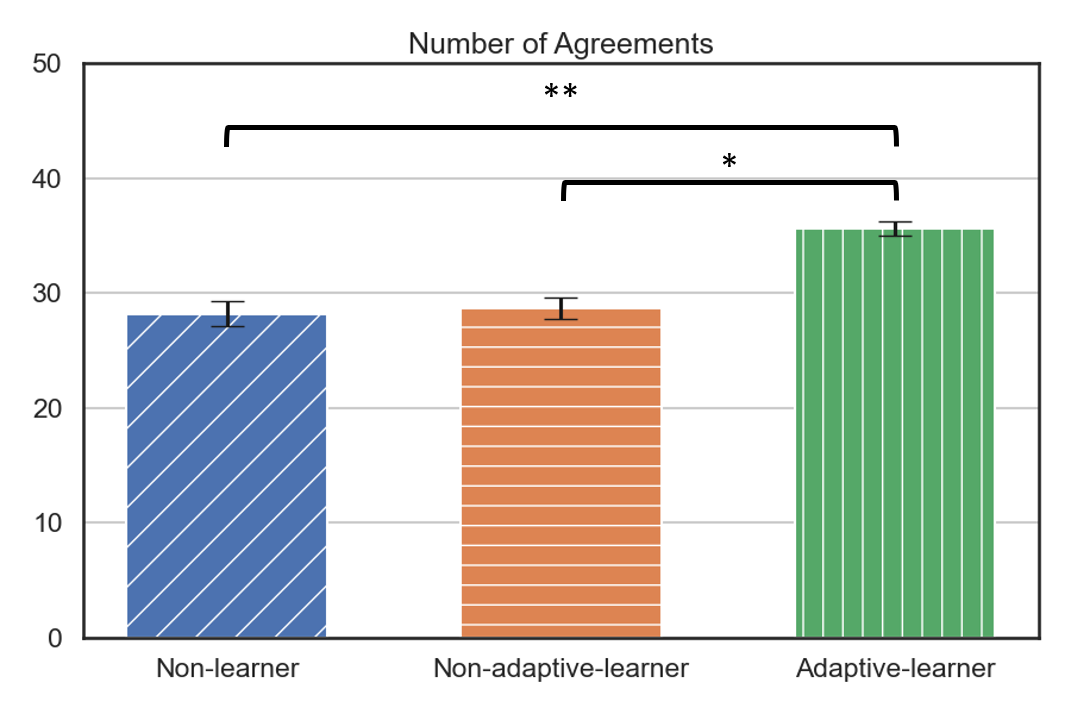}
    \caption{Exp 2-- Number of agreements }
    \label{fig:agreements-2b}
\end{figure}

Regarding the number of agreements (Fig. \ref{fig:agreements-2b}), there was a significant difference among the three strategies $(F(1.584, 36.435) = 25.829, ~p < 0.001)$.
% $(F(2, 46)=25.829,~p<0.001)$. 
Post-hoc analysis showed that there was a significant difference between the non-learner and the adaptive-learner strategies $(p<0.001)$ and between the non-adaptive-learner and adaptive-learner strategies $(p<0.001)$. 
% These differences are shown in Fig. \ref{fig:agreements-2b}
% \textcolor{red}{Jessie: @Shreyas, I moved the sentences around and couldn't find the post-hoc analysis results anymore, could you add it here?}

% and number of agreements $(p<0.001)$ and between the non-adaptive-learner and the adaptive-learner in average trust $(p=0.003)$, end-of-mission trust $(p<0.001)$, and number of agreements $(p<0.001)$, with the adaptive learner being rated the highest in all comparisons.

\begin{figure}[h]
    \centering
    \includegraphics[width=0.7\columnwidth]{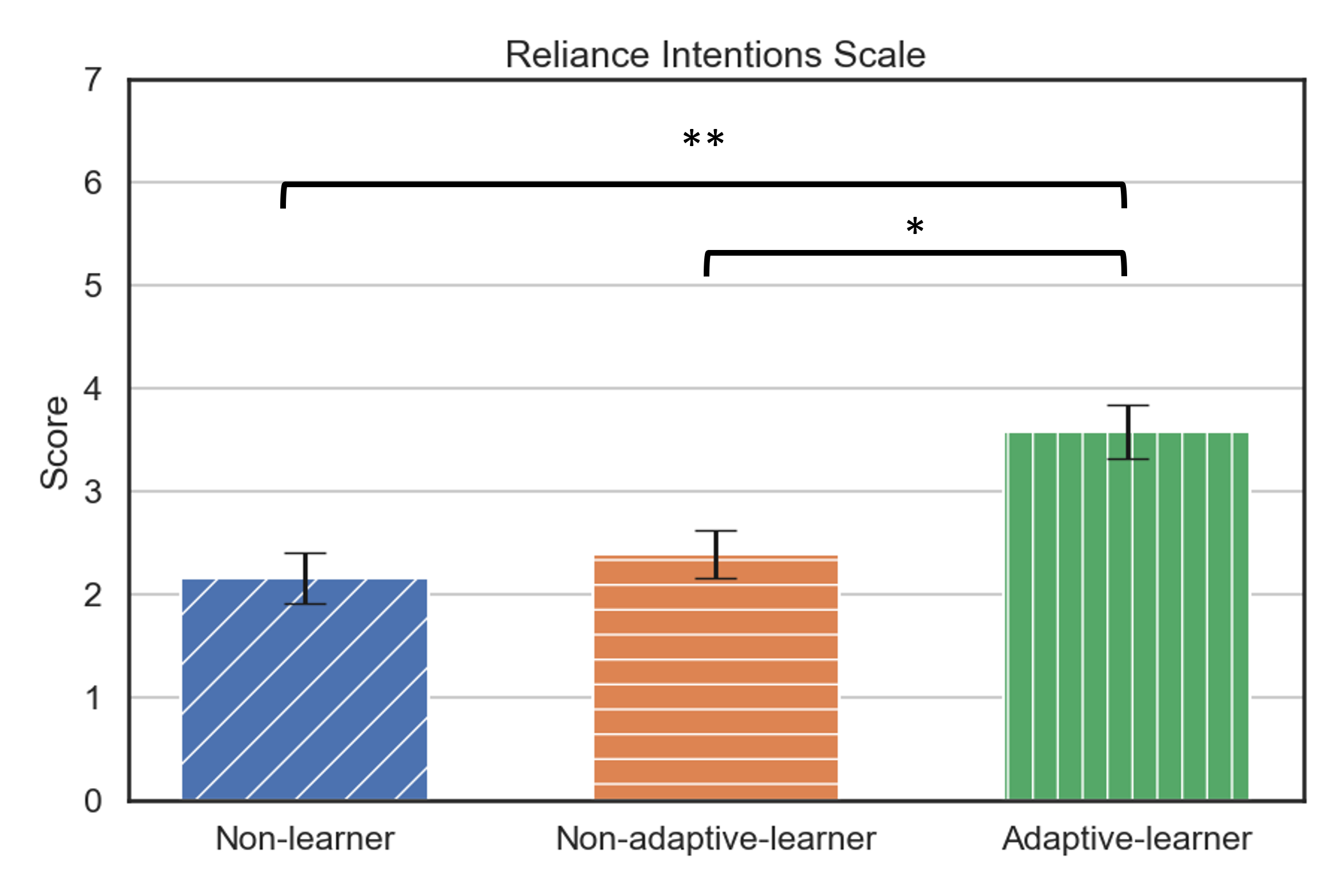}
    \caption{Exp 2-- Post-mission reliance intentions}
    \label{fig:reliance-intentions-2b}
\end{figure}

Comparing reliance intentions (Fig. \ref{fig:reliance-intentions-2b}), there was a significant difference between the three strategies ($F(2, 46) = 13.691, p < 0.001$), with the adaptive-learner strategy rated higher than the non-learner strategy ($p<0.001$) and the non-adaptive-learner strategy ($p=0.004$).

% Repeated measures ANOVA shows significant difference between the trust ratings given to the three strategies ($F(2,46) = 16.3, p < 0.001$) on Muir's trust scale, with the highest trust given to the adaptive. 

% \vspace{-15pt}

% Fig. \ref{fig:trust-muir-2b} shows the post-mission trust rating given using Muir's trust scale. 

% Fig.\ref{fig:reliance-intentions-2b} shows the post mission reliance intentions rating \cite{Lyons2020} given by the participants. 

% \newpage
\subsection{Performance}
% The performance is measured by the remaining health of the soldier and the total time to complete the mission. The results of these are presented below.

\subsubsection{Experiment 1: with informed prior}
% The performance was similar across the three strategies (Fig. \ref{fig:health-and-time-2a}). Since we were starting with an informed prior on the human's preferences, the posterior distribution for most participants was ``close" to this prior distribution, thus reducing room for any adaptation to be done by the adaptive strategy. This may be one of the reasons why the performance and trust is very similar across the three strategies. 

% Repeated measures ANOVA reveal no significant difference between the three strategies in either health at the end of the mission $(F(2,58)=1.702,~p=0.191)$ or mission completion time $(F(2,58)=1.301,~p=0.280)$.

% \begin{figure}[h]
%     \centering
%     \includegraphics[width=0.8\columnwidth]{images/barcharts/phase2a/performance/health-and-time-strategy.png}
%     \caption{The health of the soldier at the end of the mission and the mission completion time}
%     \label{fig:health-and-time-2a}
% \end{figure}

Even though there seemed to be a decreasing performance trend from non-learner to non-adaptive learner and to adaptive learner (i.e., $61.47 \pm 18.12$, $60.25\pm 17.03$, and $55.83\pm 20.39$) the trend did not reach statistical significance $(F(2,58)=2.067,~p=0.136)$. 

% \begin{figure}[h]
%     \centering
%     \includegraphics[width=0.8\columnwidth]{images/barcharts/phase2a/performance/combined-percentages.png}
%     \caption{The team performance computed as the weighted sum of percentage health remaining of the soldier and the percentage time remaining in the mission}
%     \label{fig:combined-percentage-2a}
% \end{figure}

\subsubsection{Experiment 2: with uninformed prior}
% The performance was similar across the three strategies (Fig. \ref{fig:health-and-time-2b}). This is likely because the human's action choices across the three strategies were very similar. In the case of the non-learner and non-adaptive learner strategies, this action choice was not in agreement with the recommendation, and thus, trust decreased. In the case of the adaptive learner, however, this action choice agreed with the recommendation, and thus, trust increased. Therefore, there was not any noticeable impact of the interaction strategy on the task performance. 

% Repeated measures ANOVA reveal no significant difference between the three strategies in health at the end of the mission $(F(2,58)=2.059,~p=0.139)$. However, we see a significant difference between the three strategies in mission completion time $(F(2,46)=3.538,~p=0.037)$. Post-hoc analysis with Bonferroni adjustment, however, shows no significant difference any pair of strategies. The similarity in this case can be explained through similar action choices by the human for the three strategies, irrespective of the recommendation received. Thus, the adaptive-learner strategy leads to more agreements with the recommender and higher trust, but no loss in team performance. 

% \begin{figure}[h]
%     \centering
%     \includegraphics[width=0.8\columnwidth]{images/barcharts/phase2b/performance/health-and-time-strategy.png}
%     \caption{The health of the soldier at the end of the mission and the mission completion time}
%     \label{fig:health-and-time-2b}
% \end{figure}

There seemed to be an upward trend from non-learner to non-adaptive learner and to adaptive-learner. Unfortunately, it did not reach significance. 

\subsection{Workload}

\subsubsection{Experiment 1: with informed prior}
% \textcolor{blue}{Jessie: @Shreyas, can you complete the following sentences?} 
Comparing the average workload across the three interaction strategies showed no significant difference $(F(2, 58) = 2.634, ~p = 0.089)$ (Table \ref{tab:results}). Additionally, repeated measures ANOVAs did not reveal any significant differences between the three strategies in any of the dimensions.
% \begin{figure}[h]
%     \centering
%     \includegraphics[width=0.8\columnwidth]{images/barcharts/phase2a/workload/workload.png}
%     \caption{Responses on the NASA TLX scale}
%     \label{fig:workload-2a}
% \end{figure}

\subsubsection{Experiment 2:  with uninformed prior.}
\begin{figure}[h]
    \centering
    \includegraphics[width=0.8\columnwidth]{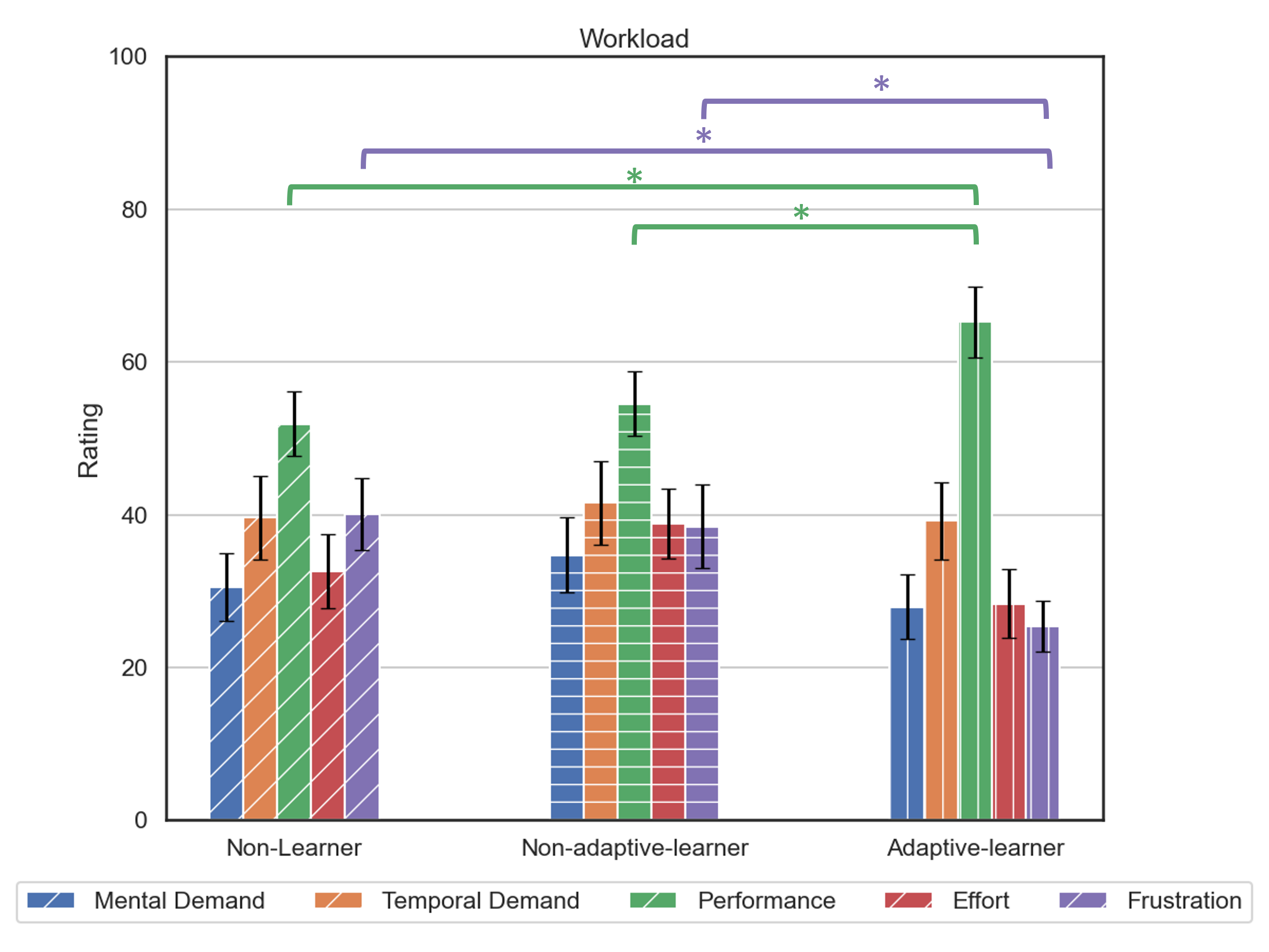}
    \caption{Exp 2: Responses on the NASA TLX scale}
    \label{fig:workload-2b}
\end{figure}

% \textcolor{blue}{Jessie: @Shreyas, can you complete the following sentences?} 
Comparing the average workload (\ref{tab:results}) across the three interaction strategies showed a significant difference $(F(2,46)=10872, ~p<0.001)$. 
Figure \ref{fig:workload-2b} shows the participants' responses on each dimension. There were significant differences between the three strategies in performance $(F(2,46)=5.443,~p=0.008)$, effort $(F(2,46)=4.252,~p=0.02)$, and frustration $(F(2,46)=5.454,~p=0.007)$. 
Pairwise comparisons with Bonferroni adjustments showed that the adaptive-learner strategy led to higher perceived performance compared to the non-adaptive learner $(p=0.037)$ and to the non-learner $(p=0.044)$ strategies and led to lower frustration compared to the non-adaptive learner $(p=0.032)$ and to the non-learner $(p=0.017)$ strategies. 

% ,  a significant difference between the non-learner and the adaptive-learner strategies in performance $(p=0.044)$ and frustration $(p=0.017)$, with the perceived performance higher and frustration lower for the adaptive learner. 
% There is also a significant difference between the non-adaptive-learner and adaptive-learner strategies in performance $(p=0.037)$ and frustration $(p=0.032)$ with the perceived performance higher and frustration lower for the adaptive learner.

\section{Discussion and Conclusion}
\label{sec:conclusion}

In this study, we developed and compared three robot interaction strategies: the non-learner strategy, the non-adaptive-learner strategy, and the adaptive-learner strategy, with and without an informed prior for the IRL learning algorithm. We focused on evaluating their influence on various human trust in the robot, agreement, reliance, workload, and team performance.
% dependent variables, including human trust in the robot, agreement, reliance, and workload, as well as their impact on team performance. 

% \subsection{Value Alignment with (Accurate) Informed Prior}
\subsection{Value Alignment with an Informed Prior}
One critical insight from the research is the observed ``uniformity'' across all three interaction strategies when the IRL algorithm is initiated with an informed prior. In Experiment 1, the informed prior was calculated by training on a previously collected dataset using the same testbed and with participants from the same population as this study. Therefore, the calculated prior was considered an accurate estimation of the participants' true values/goals. The resulting prior was realistically skewed toward saving health rather than saving time. With this accurate informed prior, the three interaction strategies were more-or-less indistinguishable. Among all the dependent measures, including trust, agreement, reliance intentions, performance, and workload, we observed only one significant difference in reliance intentions. This large uniformity across the three strategies could be because, with an accurate informed prior on human values, there is no room for significant alignment to be done by the adaptive-learner strategy. This highlights that when a robot's reward function is already closely aligned with that of the human \textit{a-priori}, adaptive strategies may exhibit negligible benefits.

Out of expectation, we observe a significant difference in post-mission reliance intentions, that participants were willing to rely on the non-learner more than the adaptive-learner strategies. This result could have been because any alignment from the adaptive-learner strategy, although very limited, could be regarded as a lack of 
% consistency or 
predictability. As pointed out in research on teamwork, the ability to predict the actions of one's teammate is vital \cite{cooke_interactive_nodate, klien_ten_2004}. 

% \textcolor{orange}{Jessie: @Shreyas, do you have any explanation for Figure 3?}
% \textcolor{blue}{Shreyas: @Jessie, I am not sure. Looking closely at the data, the difference mostly comes from Q3, Q4, and Q6 from the questionnaire}
% \textcolor{blue}{
% \begin{itemize}
%     \item Q3 - I would feel comfortable relying on the agent in the future.
%     \item Q4 - If I had my way, I would NOT let the agent have any influence over issues that are important to the task (search and rescue).
%     \item Q6 - I would be comfortable allowing the agent to implement its search decision, even if I could not monitor it.
% \end{itemize}
% With the adaptive-learner being rated the lowest, the non-adaptive-learner in the middle (non-significant) and the non-learner the highest. 
% I am not sure why this difference is arising even though the behavior and interaction with the three strategies is more-or-less the same.}

\subsection{Value Alignment with an Uninformed Prior}

In Experiment 2, when the IRL learning algorithm was initiated with an uninformed uniform prior, we observed significant benefits of value alignment. The adaptive-learner strategy led to significantly higher trust, high agreement, and reliance intentions. In addition, participants had higher perceived performance and lower frustration interacting with the adaptive learner, while the team performance was maintained. Thus, in the absence of a good initial prior, our adaptive framework can be used to build trust in the robot. This scenario underlines the importance of adaptive strategies in real life where a good \textit{a-priori} estimation of the human's values/goals is oftentimes unavailable, offering an approach to building trust while maintaining performance.

The results from Milli et al. [30] showed that robots should not be completely obedient to humans who are not acting rationally. Instead, when interacting with such humans, the robot should use its estimation of the human's underlying reward function. We extend this by providing a human-subjects study showing that personalized value alignment is only beneficial when a good prior on the human's reward weights is unavailable

\subsection{Implications for a broader HRI context}

The algorithmic focus of existing work on value alignment \cite{Hadfield-Menell2016, Fisac2020, Yuan2022} has one implicit assumption: aligning a robot’s values with a human’s is beneficial. Our study is the first attempt to examine whether and to what extent such alignment can benefit trust, workload, and team performance. We show that personalized value alignment is beneficial only when an informed prior is unavailable.
% only when an informed prior is unavailable, personalized value alignment is beneficial. 
% However, when starting with a good informed prior, the benefits diminish. 
The implications of our study should extend 
% beyond the specific context of our research—a reconnaissance mission scenario—
to other real-life HRI scenarios involving conflicting objectives. For instance, a rehabilitation robot must balance a patient’s pain tolerance with long-term health goals when assigning the appropriate level of exercise. 
% Furthermore, our findings are crucial for the integration of commercial robots into domestic environments. 

Our study involved a relatively homogeneous participant group, leading to the calculation of a fairly accurate informed prior. However, achieving such accuracy in real-world settings with demographically diverse individuals is more challenging. In such cases, aligning a robot's values with those of individual human users becomes essential. The benefits obtained from value alignment are key for the acceptance and adoption of robots in homes and workplaces, highlighting the need for adaptable strategies in HRI design. 

Further, the idea of incorporating a layer of trust in the decision-making system of an intelligent agent trying to align its values to that of the human user is an interesting area to explore in other HRI domains like shared control, social robotics, etc.

\subsection{Limitations and Future Research}
The results of our study should be seen in light of the following limitations. First, we provide a demonstration in the case when there are only two components in the team's reward function. Therefore, we only need to learn the human's preference for one of the two components and can ascertain their relative preference between the two objectives. Our formulation, however, can readily be extended to the case where there are more than two objectives in the team's reward function, with additional computations required to learn and maintain a distribution over each reward weight. 

Second, our simulated scenario consists of binary actions.
% : to use or not use the armored robot. 
Judging the performance of the recommendations is fairly easy in this case, since we only need to compare the rewards earned for these two actions. In case more than two actions are available, this assessment becomes more difficult. Thus, although the human trust-behavior model can be readily extended to such a case of multiple actions, extending the trust dynamics model is challenging and is an interesting avenue for future research.

Third, in our scenario, there is an expected skewness among the general population to be more concerned about saving health. 
% Thus, the data-driven prior was a good representation of the population's values/goals. 
It would be interesting to study cases where the two objectives are more balanced, resulting in a more balanced informed prior. In such cases, personalized adaptation may still be beneficial.
% \textcolor{blue}{

Finally, our scenario, which entails a trade-off between "saving health" and "saving time", and the decision to use or not use an armored robot, is informed by the complex decision-making scenarios in real-life HRI contexts, such as DARPA’s SQUAD-X program in which individuals receive recommendations from air and ground robots for various tasks. While our scenario offers insights into these types of decisions, we recognize it as a simplified representation of situations where decisions involve numerous objectives, a variety of recommendations, and possible actions. Therefore, further research is essential to determine the applicability of our findings in more complex, real-world environments and to validate the robustness of our conclusions in diverse and dynamic HRI settings.
\begin{acks}
This work was supported by 
% [Anonymity].
the Air Force Office of Scientific Research under Grant FA9550-20-1-0406.
\end{acks}

\bibliographystyle{ACM-Reference-Format}
\bibliography{sample-base}

% \appendix









\end{document}